\documentclass[conference,a4paper]{ieeetran}

\usepackage[utf8]{inputenc}
\usepackage{lmodern}
\usepackage{textcomp}
\usepackage{subeqnarray}
\usepackage{amsfonts}
\usepackage{amsmath}
\usepackage{amssymb}
\usepackage{graphicx}
\usepackage{color}
\usepackage[colorinlistoftodos]{todonotes}
\usepackage{fancyvrb}
\usepackage{verbatim}
\usepackage{latexsym}
\usepackage{multicol}
\usepackage{soul}

\title{Deep Learning for automatic sale receipt understanding}

\author{\authorblockN{Rizlene Raoui-Outach\authorrefmark{1,2}, Cecile Million-Rousseau\authorrefmark{1} ,Alexandre Benoit\authorrefmark{2} and Patrick Lambert\authorrefmark{2} }
\authorblockA{\authorrefmark{1} AboutGoods Company Annecy France\\
e-mail: \emph{\{rizlene.raoui, cmr\}@aboutgoods.net}}
\authorblockA{\authorrefmark{2} Univ. Savoie Mont Blanc, LISTIC, F-74000 Annecy, France\\
e-mail: \emph{\{alexandre.benoit, patrick.lambert\}@univ-smb.fr}}}

\begin{document}
\IEEEoverridecommandlockouts\pubid{\makebox[\columnwidth]{978-1-5386-1842-4/17/\$31.00~\copyright{}2017 IEEE \hfill} \hspace{\columnsep}\makebox[\columnwidth]{ }}
\maketitle
\begin{abstract}
As a general rule, data analytics are now mandatory for companies. Scanned document analysis brings additional challenges introduced by paper damages and scanning quality.
In an industrial context, this work focuses on the automatic understanding of sale receipts which enable access to essential and accurate consumption statistics. Given an image acquired with a smart-phone, the proposed work mainly focuses on the first steps of the full tool chain which aims at providing essential information such as the store brand, purchased products and related prices with the highest possible confidence. 
To get this high confidence level, even if scanning is not perfectly controlled, we propose a double check processing tool-chain using Deep Convolutional Neural Networks (DCNNs) on one hand and more classical image and text processings on another hand.
The originality of this work relates in this double check processing and in the joint use of DCNNs for different applications and text analysis.

\end{abstract}

\begin{keywords}
    Receipt image understanding, Deep Convolutional Neural Networks, Object Detection, Semantic Analysis
\end{keywords}

\section{Introduction}
In the field of mass distribution, the knowledge of consumer behaviors is a key data that many companies are looking for. Indeed, such information has a high added value as it provides accurate consumption statistics. These statistics are an important input for different studies which aim to develop effective sales strategies. 
Currently, such data is manually obtained by recruiting consumers as panelists who are asked to scan the purchased products and to fill out forms. Such a solution is costly and thus cannot be applied on large populations which limits its statistical value and significance. 
Reading a sale receipt is also of real interest for specific cases such as discount coupon awarding validation. Consequently the automatic understanding of sale receipts is very challenging.
First of all information retrieval from such document is not that easy because receipts are often damaged before being scanned and because the textual content consists of a non-standardized terminology. This work has been developed within the collaboration between a university research lab (LISTIC) and the AboutGoods start-up. 
This company is specialized in the relationship between consumers and distributors investigating mobile and Internet solutions to improve communication between all stakeholders. 
For both reducing costs and enlarging panelist community, 
AboutGoods developps a free mobile application that enables any consumer to take and send a photo of its receipts. Then, an online framework decodes the receipt image and performs a data analysis. The aim is to offer assistance to consumers for managing their budget. 
Such image analysis problem is not straightforward and cannot be solved with a single OCR (Optical Character Recognition) step. Indeed the life of a consumer receipt is highly subject to degradations such as crumples, tears and other physical damages before being sub-optimally captured using a smartphone. Furthermore, a strong industrial constraint relating to the reliability of each extracted information is imposed. Each piece of information must then be extracted by competing algorithms which results are merged to maximize reliability and to favor false detection rather than no detection since our industrial constraint does not allow any receipt to be missed.
It results in a multi-step processing chain as described in (Fig.\ref{fig:wholeProcessing}).\\
The rest of the paper is organized as follows. In section 2, we present related works. Section 3 provides the description of our approach. 
Performance analysis is presented in section 4, and we conclude in section 5.

\section{Related works}

There are only a very few works dealing with sale receipt analysis from picture acquired with a smartphone. 
Typical approaches rely on a confident image acquisition such as \cite{SpectClust:ReceiptIdentif} which facilitates character recognition. So the following state of the art will rely on the three main steps, not dedicated, but necessary for receipt analysis, i.e object detection, character recognition and semantic analysis, with a special emphasis on object detection which is the core of the proposed work.

\subsection{Object Detection}
In our application, an object can be a ticket, a text block or a logo. In the literature, there is a lot of works dealing with the detection of such objects.

A first group of conventional works relies on local feature detection, description and matching methods using engineered features such as SIFT or HOG. Augereau \& al \cite{AdaptativeSIFT:Augereau} proposed a solution to localize semi-structured documents such as ID cards or tickets. However, in the case of document images mainly containing texts, interest point matching is disturbed by the repetitive character patterns. They therefore refine this approach by introducing a relevant key point selection method and a specific implementation of the RANSAC algorithm that is made robust against character redundancy. But the proposed strategy does not satisfy enough our high confidence information extraction constraint in case of noisy images.
Using also a classical approach, a set of works localizes text areas within an image by applying region detection and classification \cite{TextLocalizationLJ:withStroke}. These approaches, mainly dedicated to complex natural scene analysis, are too sophisticated for our problem.
Engineered features are also widely used for logo detection as in \cite{SIFT:multiScaleRank} and proved to be efficient. However, once more, the frequent presence of characters in store logos causes these approaches to fail in many cases.

State of the art recently highlighted a second category of approaches, based on Deep Learning (DL), which outperformed all the previous approaches on many object detection tasks in images \cite{AlexNet:alexNet2012,GoogLeNet:GoogLeNet2015}. The enabling factor is however the availability of large annotated datasets in order to optimize a huge amount of parameters. A classical solution is then to fine-tune pretrained networks as explained in \cite{FineTuningDL}. Our work relies on this kind of solution since it enables higher confidence levels.\\
Concerning text block detection, DL based methods show promising results such as \cite{Deep:textSpotting} that relies on a multi-task network and allows words spotting and recognition in uncontrolled scenes. In \cite{Moysset:2015:PTS:2880452.2880785} the authors propose an approach for text line localization based on Convolutional Neural Networks and Multidimensional Long Short-Term Memory cells. In contrast, we process text within specific regions or text blocks since they present different semantic contents (header, list of products, prices...).\\
Regarding logo detection, some recent works have dealt with localization and classification of logo using DL. For instance, in \cite{bianco2015logo} the proposed method first selects candidate subwindows using an unsupervised segmentation algorithm, and then used an SVM-based classification of such candidate regions using features computed by a pre-trained Convolutional Neural Network (AlexNet \cite{AlexNet:alexNet2012}). \cite{DeepLogo:Detection17} has also used pre-trained deep networks for logo detection. Obtained results are interesting and we also consider such an approach, but, as logos are sometimes only characters, the detection has to use jointly character recognition.
\vspace{-0.15cm}
\subsection{OCR}
\vspace{-0.15cm}
Optical Character Recognition (OCR) has been a very active field for many years. More specifically, many applications addressing printed characters recognition already achieve good performance levels. Tesseract \cite{reconnaissance:Tesseract}, ABBYY FineReader\footnote{https://www.abbyy.com/en-eu/} are widely used. In our specific application, they do not achieve perfect text recognition. Recently, the Google Vision API\footnote{https://cloud.google.com/vision} proved to be one of the most efficient OCR but remains as a blackbox and does not ensure all the text to be recognized. We however use this as the best compromise. In addition, since the focus of this paper relates to the first steps of the processing chain, we do not need to go further into the details of the OCR topic and use it as a black-box tool.

\vspace{-0.1cm}
\subsection{Semantic analysis}
\vspace{-0.1cm}
OCR results can be improved by using high level complementary analysis. Exploiting character recognition confidence levels, using dictionaries, etc. are classical solutions in this way. When dealing with a specific domain, a higher step consists in building and using a specific ontology \cite{semantique:ontology}. The ontological approach is essential for the understanding of domain concepts and particularly of product concepts. As for OCR, this topic is out of the scope of the paper and we use it as a tool to finalize the processing chain.

\section{Proposed approach}

\subsection{Global description}

\begin{figure*}
  \includegraphics[width=18cm]{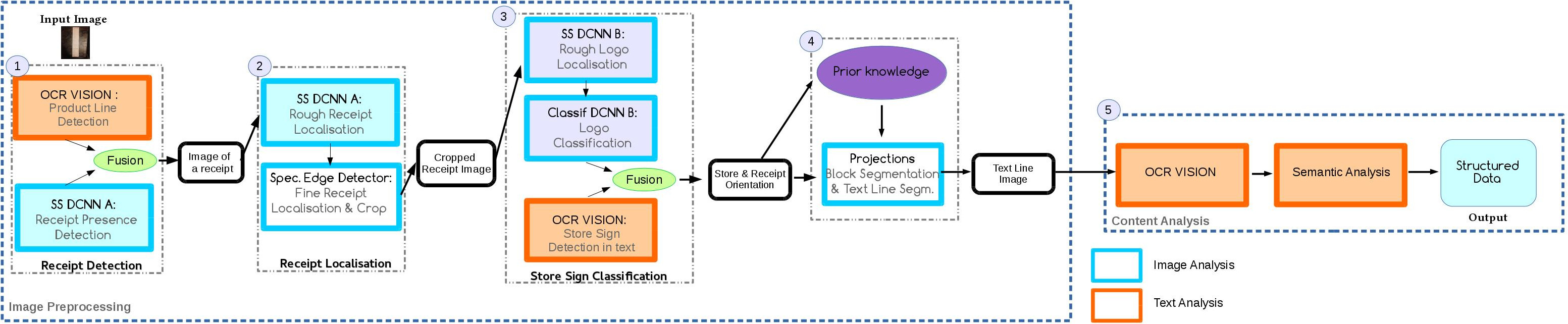}
  \caption{Complete chain of an automatic reading system of sale receipt}
  \label{fig:wholeProcessing}
 \vspace{-0.5cm}
\end{figure*}

The global tool-chain is presented in Fig. \ref{fig:wholeProcessing}. A first important stage, which can be considered as pre-processing, is composed of several steps : the receipt detection (is there a receipt in the image?), the receipt localization and cropping, the store brand logo detection and classification and finally the text block detection. Those steps, which are the core of this paper, are mandatory to facilitate the following higher level processing.
Receipt detection is necessary because users do not always follow the rules and upload blurred images to the servers, selfies, etc. that must be detected not to fool our system. 
Then, a receipt must be localized and cropped to eliminate background which can be regarded as noisy information. The next step is the store brand logo localization and classification. Indeed, knowing the store brand allows prior knowledge about the structure and organization of the sale receipt to be introduced and facilitate further processing.
The last preprocessing step consists in text block segmentation in order to separate the header, the list of products, prices etc. This step enables both accurate content interpretation and OCR errors to be fixed in a refined way. For instance, "I00' in the price list is likely to be "100".

Once the pre-processing stage is achieved, each text block is submitted to an OCR. 
Finally, a semantic text analysis allows us to extract essential information. Semantic text analysis based on an ontology \cite{semantique:ontology} is able to automatically interpret short product labels into a full and unique product label, which is a difficult task as there is no standard denominations. These last steps will not be detailed more since the paper focuses on the pre-processing.

\subsection{Overview on the used DCNNs}

For receipt pre-processing, Deep Convolutionnal Neural Networks (DCNNs) are used at different steps and are of two types : the first one is dedicated to classification while the second one is dedicated to semantic segmentation. However, the first layers of both networks rely on the same architecture and learned weights. Only the last layers are specialized for their specific tasks (transfer learning).

\textbf{Classification Network (C DCNN)}

Classification is a common task performed by DCNNs. However, this task generally requires a training phase using a huge amount of annotated data. Such prior condition cannot be fulfilled in our context. 
As a result, we have resorted to the use of pre-trained networks which have good performance level in similar situations. Two well-known Deep Learning networks, AlexNet \cite{AlexNet:alexNet2012} and GoogLeNet \cite{GoogLeNet:GoogLeNet2015}, pre-trained on the 1000 class problem of the ImageNet 2012 dataset \cite{ImageNet:Dataset}, were tested. 
After transferring these two networks, experiments on our datasets and related tasks showed that AlexNet and GoogLeNet have close performances. But GoogLeNet involves much less parameters.
Here, transferring consists in training only the last layer on our dataset after replacing the 1000 neuron last layer by a new one with randomized weight values and with a number of neurons corresponding to the number of classes of our application.
 
\textbf{Semantic Segmentation Network (SS DCNN *)}

DCNNs can also be used for pixel classification, denoted semantic segmentation. In our context, this allows the receipt and its store sign to be localized. This segmentation network is obtained by transforming the classification network described above into a fully convolutional network. In this way, a fixed size sliding window is applied on the whole image with a stride corresponding to the network field of view. The result is a set of sub-sampled images (one for each target class), each image being a probability heat map of belonging to a target class. The sub-sampled image resolution remains consistent with respect to our final aim. 
To do so and considering the GoogLeNet architecture, the last fully connected layer (input size is N$\times$1024$\times$1$\times$1, N being the number of outputs) is transformed into a convolutional layer with N kernels of size  1024$\times$1$\times$1. 

\subsection{Receipt or not receipt?}

The first pre-processing step is to automatically decide whether there is a sale receipt in the image or not. As we can see in Fig.\ref{fig:wholeProcessing}, at the first step, two different approaches are considered in order to avoid false negatives and ensure the reliability of the automatic processing.

\textbf{Text analysis for receipt detection}

The first step consists in a text analysis using an OCR applied on the whole image.
At this level, OCR does not work well because of the noisy information provided by the background. However even a rough result is enough to validate the presence of a receipt within the image.  Technically, a receipt is detected if the text provided by the OCR contains at least one ``product line". A product line is a sequence of characters corresponding to a very specific string format: typically a set of letters, figures, spaces and punctuations on the first part, followed by spaces and ending with a string respecting a price format (figures possibly with a comma or a point and possibly a monetary symbol) (Example : ``BRICK LP \hspace{2cm}0.79€"). This detection is performed by using regular expression formalism.

\textbf{Deep Learning analysis for receipt detection}

The image analysis based approach relies on the semantic segmentation network "SS DCNN A" (\ref{fig:wholeProcessing}). It was trained on a two class problem, namely, "receipt" or "not receipt".
Once the network has processed an image, the resulting heat map consists in a gray level image where the higher the pixel intensity, the greater the probability that this pixel corresponds to a receipt. This gray level image is thresholded with a value (70\% of the full scale) that has been experimentally defined thanks to tests performed on 500 annotated images. 
This value satisfies the no missed receipt constraint. Therefore, an input image is regarded as containing a receipt if the thresholded image has at least 25\% positive (white) pixels. This experimental threshold is voluntarily low to ensure that our detector takes into account very long receipts that occupy a small surface of the image. As a comparison, experiments showed that the positive pixels rate of ``no receipt" images does not exceed 10\%. Once again, this fulfills our no missed receipt constraint. 

\textbf{Fusion of textual and image results}

In order to minimize further the missing rate while enhancing detection confidence, the two competing and complementary methods based on text and image analysis are fused using the logical OR operator.

\subsection{Receipt localization and crop}

\begin{figure}
  \includegraphics[width=8.7cm]{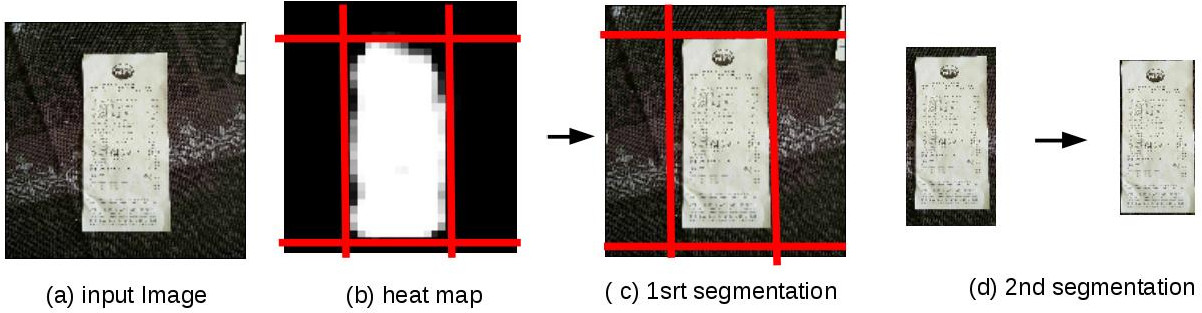}
  \caption{Receipt Detection and Localization}
  \label{fig:receiptDetectionProcess}
  \vspace{-0.3cm}
\end{figure}

Once all the "no receipt images" have been filtered out, remaining receipt bounding box areas must be cropped (see Fig.\ref{fig:wholeProcessing}, step 2) before moving to the next steps.

First the heat map provided by the "SS DCNN A" used in the previous step allows to get a rough and wide localization of the receipt and enables a first wide crop. More into details, the 70\% heat map thresholding described in the previous section is reused here in order to identify the receipt area to crop. Resulting crop is always wider than the receipt, so a refined cropping is applied as described on Fig \ref{fig:receiptDetectionProcess} (c).

The refined cropping method consists in the use of a specifically designed contour detectors proposed in \cite{ReceiptDetection:RFIA} that detects points belonging to the sale receipt edges. The basic principle of this receipt edge detector is to look for regions with two main adjacent areas, a bright one (the ticket border) and a darker one (the background), in quasi vertical or horizontal position. Once the 4 edges are obtained, simple geometric considerations enable the final crop of the sale receipt (Fig.\ref{fig:receiptDetectionProcess}(d)).

\subsection{Store sign recognition}

As for receipt detection, store sign recognition involves two different approaches (once again, image and text analysis), as we can see in Fig.\ref{fig:wholeProcessing}, step 3. The aim is to ensure the reliability of the extracted information, both of them bringing complementary information. Image analysis also enables receipt orientation detection (top/bottom inversion) assuming that the store sign is located at the top of the document. In addition, sign localization facilitates the text block segmentation step. Indeed, as sale receipts are structured documents, once the store sign is known, one can use prior knowledge on the brand to better understand the receipt structure.

\textbf{Semantic text analysis for store sign recognition}

Three criteria are proposed to ensure reliable recognition of the stores from the OCR output. The first criterion is based on the store name. Having a homemade store name database, a known store name is searched within the decoded receipt text. N-gram with N=2 is considered to compensate for OCR errors from unrecognized characters to added spaces. In the end, the most confident store name is kept.

The second criterion consists in identifying the presence of a phone number using a regular expression. If detected, this phone number is compared to the ones of a known store brand phone number database.

Finally, the third criterion is based on references to the terminology of the sign. For each sign, the terminology used in terms of slogan, loyalty program and brand distributor have been listed by the company. The terms being specific to each sign, finding such a term makes possible the sign identification.

The final result is a list of pairs whose first element is the identified store sign and the second element is a weight that represents the number of criteria that identified the sign. This list may be empty, especially when the sale receipt is not fully captured in the image. 

\textbf{Logo detection and store sign recognition}

\begin{figure}
\includegraphics[width=8cm]{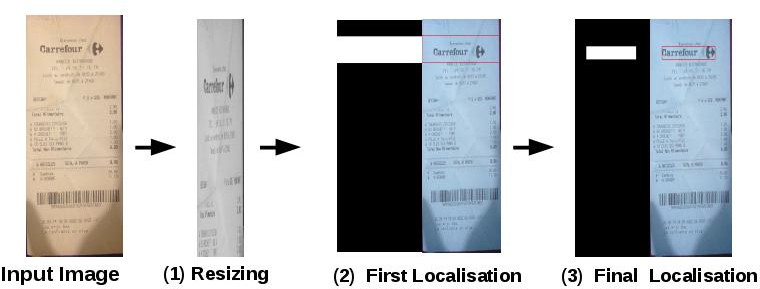}
  \caption{Logo Localization and classification process}
  \label{fig:logoDetectionProcessing}
   \vspace{-0.3cm}
\end{figure}

In order to recognize the store sign logo printed on each receipt, we use both the classification and the semantic segmentation DCNNs.
Among all the store logos considered in this work, one can roughly observe that there are two types of logos, all being rectangular with the widest sides on the horizontal axis: the long ones and the short ones. This allows two main height/width ratios to be defined and the logos to be resized taking them into account so that at least one version will generate a squared logo that will fit into a 227x227 pixels image to feed into the deep network. Fig.\ref{fig:logoDetectionProcessing} shows an illustration of an appropriate resizing at step 1. 
One must also mention that the two resize factors are chosen in order to ensure that the sliding window generated by the SS DCNN will contain, at least once, the full logo. 

From a practical point of view, to save time and as the long logos are more frequent, we first use the resize ratio corresponding to this type of logos. If the semantic segmentation network (SS DCNN B, step 2) does not detect logo area, the second ratio factor is applied. Potential sign areas are next cropped and classified by the "C DCNN B"
The only retained area is the one that obtains the highest store sign classification probability. 

Going further, since the logo is located at the top of the receipt, this strategy is applied only on the upper part of receipt images. Only if no logo is detected, then the lower part is processed. This strategy allows both computation cost reduction and receipt orientation detection.
As a counterpart, at this stage, the confidence in the precise store sign classification is not sufficient. Indeed, the coarse resolution of the semantic segmentation leads to a wide crop around the sign area that may include distracting neighboring information (white borders or text) that may impact on the brand  classification accuracy.

Then a last step is necessary to refine this preliminary classification result. In order to remove the neighboring information, the wide logo crop is binarized and a pixel projection, as described in \ref{subSec:TextLocalization}, allows to keep only text areas without white areas. 
As many text blocks can be generated, each one is once again classified by the "C. DCNN B" and the best result is retained. This two-step operation brings about a significant increase in performance. 

\vspace{0.2cm}
\textbf{Store sign recognition}
\label{subsubsec:signrecognitionRules}

The last step of the process consists in fusing the DCNNs based results with the semantic text analysis results. The name of the store sign is automatically associated to the sale receipt in one of the two following cases: 

\begin{itemize}
\item The 3 semantic text analysis criteria provide the same store sign. In that case image analysis is not used;
\item The store sign provided by the DCNNs matches the store sign provided by at least 2 semantic analysis text criteria.
\end{itemize}
In other cases, automatic processing is too uncertain and the related receipts are redirected to a human expert analysis.

\subsection{Text block localization}
\label{subSec:TextLocalization}

Once the store sign is recognized, the following steps focus on the processing of the remaining textual information. First, text blocks must be detected (Fig.\ref{fig:wholeProcessing}: step 4) before OCR and semantic analysis. This step is quite easy as there is a good contrast between the characters (dark pixels) and the background (bright pixels). So an adaptive binarization is used. Then blocks covering the entire width of the receipt are extracted in a classical way using horizontal pixel intensity projections. These blocks are then divided into sub-blocks using vertical projections in order to separate products labels and their prices for example. Since the store sign is known, prior knowledge is used to adapt block boundary splits. Finally, on each sub-block, a text line segmentation is carried out before being submitted to the OCR. It can be noted that the projection approach may require a prior rotation of the receipt which can be easily performed as the receipt borders have already been detected.

\subsection{Text semantic analysis}
Among all the information contained in a receipt, the purchased items list is the most important. Text analysis on the related receipt image blocks is then performed in step 5 of the whole process depicted in Fig.\ref{fig:wholeProcessing}. As said, this block contains short item labels which are often vague even for the consumer himself. No encoding standards exist and they often differ from one store sign to another and even between stores of the same brand for the same product.
In our industrial context, the challenge is to be able to identify precisely each of the purchased products. To this end, a dedicated product ontology has been built, the most specific concept being products grouped into product categories. Each concept has a list of terms denoting it. Currently, more than 4 000 concepts have been considered avoiding product variants in terms of trademark, packaging, quantity, etc.\\

An algorithm dedicated to text semantic analysis was then developed to associate a given short label from the OCR extraction and the product concept in the ontology. This work is still ongoing and  is not detailed as it is out of the scope of this paper.

\section{Results}

The approach is validated on a set composed of 5000 images (3000 receipts and 2000 non receipts) on which each step of the proposed pipeline is evaluated. Unfortunately no available baseline allows us to compare with. Experiments have been made possible thanks to the MUST computing center of the University of Savoie Mont Blanc.

\subsection{Receipt or no receipt ?}
\label{subSec:TicketDetectionResult}

This first step relies on a SS DCNN A based on GoogLeNet architecture pretrained on ImageNet and fine tuned on a two class problem, namely,``receipt" or ``not receipt". To do so, the pretrained architecture weights remain constant and a new last layer is considered and trained with 3500 images manually annotated. This dataset is different from the validation set and all images are real captures sent by end-users from the AboutGoods mobile application. 2400 images are receipt images while the remaining 1100 are not. In order to increase the number of training data, 10 randomly translated crops of size 227x227 pixels are randomly extracted from each image. Excellent performance levels are obtained as shown in tab.\ref{tab:ReceiptDetectionPerformances}. The fusion between the text and image approach improves overall performance as error cases are different. Text semantic can miss a receipt if the picture is slightly blurred or if the OCR doesn't recognize text whereas DCNNs can miss a receipt when brightness contrast is low.

\begin{table}
\begin {center}
\begin{tabular}{|c|c|c|c|}
\hline
	\multicolumn{4}{|c|}{\textbf{Precision Measure}}\\
    \hline
	\textbf{Class} & \textbf{DCNN} & \textbf{Text Sem.}  & \textbf{DCNN \& Text Sem.} \\
    \hline
    Receipt & 98.5\% & 98.7\% & 99.6\% \\
   \hline
   Not Receipt & 90\% & 67\% & 93\% \\
   \hline
   \hline
   \multicolumn{4}{|c|}{\textbf{Recall Measure}}\\
    \hline
   \textbf{Class} & \textbf{DCNN} & \textbf{Text Sem.}  & \textbf{DCNN \& Text Sem.} \\
    \hline
    Receipt & 99.5\% & 92.8\% & 99.9\% \\
   \hline
   Not Receipt & 91.3\% & 91.07\% & 96.8\% \\
   \hline
\end{tabular}
\vspace{-0.3cm}
\caption{Receipt Detection performances}
\label{tab:ReceiptDetectionPerformances}
\end {center}
\vspace{-0.9cm}
\end{table}

\subsection{Receipt localization and crop}
Regarding receipt localization and crop, we evaluate localization performance using the Intersection Over Union metric (IoU). In tab.\ref{tab:ReceiptLocalizationPerformances}, results show the great improvement provided by the combination of the two methods. Indeed, the first crop enabled by the DCNN approach allows the performances of the specific precise receipt detector to be improved since it eliminates many false detections generated by textured background.

\begin{table}
\begin{center}
\begin{tabular}{|c|c|c|}
\hline
\multicolumn{3}{|c|}{\textbf{Receipt Localization and Crop - IoU}}\\
\hline
\textbf{Detector Only} & \textbf{Deep Only} & \textbf{Detector \& Deep}\\
\hline
0.75 & 0.71 & 0.86  \\
\hline
\end{tabular}
\vspace{-0.2cm}
\caption{Receipt Localization performances}
\label{tab:ReceiptLocalizationPerformances}
\end{center}
\vspace{-0.9cm}
\end{table}
\subsection{Store sign recognition}

Experiments are performed with 17 different brand logos. Considering the approach detailed in section \ref{subsubsec:signrecognitionRules}, we train GoogLeNet network as described in Section \ref{subSec:TicketDetectionResult} with an equilibrated set of 1530 logo images, also different from the validation dataset, resized to size 227*227 that matches the DCNN input and ensure accurate training.
A first part of this dataset is used for the training phase with 60 images per class and a second part is considered for the test phase with 30 images per class.\\
The effect of the fusion of the image and text analysis is reported in tab.\ref{tab:LogoDetectionPerformances}. An accuracy of \textbf{98,7\%} is then reached and outperforms the accuracy of each method taken separately.
Regarding the DCNNs approach, we find that the small logos and those which are located close to the receipt edges are the most difficult cases that impact performance, because those situations are rare samples in the training dataset. 
Semantic text analysis is strongly impacted by the OCR quality so that results are not always trustworthy. Also, some specific receipts provide no textual information enabling store sign recognition. 

\begin{table}
\begin{center}
\begin{tabular}{|c|c|c|c|}
\hline
\multicolumn{4}{|c|}{\textbf{Store Sign Recognition}}\\
\hline
\multicolumn{2}{|c|}{\textbf{DCNN}} & \textbf{Text Sem.} & \textbf{DCNN and Text Sem.}\\
\hline
\textbf{Top-1} & \textbf{Top-2} & \textbf{Top-1} & \textbf{Top-1}\\
\hline
89.9\% & 95.8\% & 81.4\% & 98.7\% \\
\hline
\end{tabular}
\vspace{-0.2cm}
\caption{Logo Detection performances}
\label{tab:LogoDetectionPerformances}
\end{center}
\vspace{-0.9cm}
\end{table}

\subsection{Semantic text analysis}

The 3000 receipts of the dataset have been processed to detect the purchased products which amounts for more than 87,000 short labels.
Accurate association rate of short-label and product-concept is 81,7 \%. 
Errors are reported when (1) short labels spelling mistakes occur if the OCR doesn't correctly recognizes all the characters. (2) short labels have no corresponding concept in our ontology. Continuous ontology enrichment will then overcome this last problem.

\section{Conclusion and Future Works}

This paper presents a solution for sale receipt analysis in order to build reliable consumer analytics. From an image acquired with a smart phone, the proposed approach enables to filter out non receipt images, to localize and classify the brand logo, to localize text block and to interpret essential information from the extracted text. Special care is given to the reliability of each piece of extracted information. Thus, whenever possible, competing and complementary approaches are considered and fused. On one side, we take advantage of Deep Neural Networks trained for classification and semantic segmentation for a better understanding of the image content. On the other side, we use text semantic analysis based on ontology and linguistic in order to interact with an OCR by fixing its errors and enabling the automatic interpretation of short product labels into a full and unique product label.\\
Further work will address semantic text block segmentation using Deep Neural Networks in order to automatically classify the different text blocks (i.e. the header, product short labels etc..).
Ontology enrichment will also be considered.

\bibliographystyle{./IEEEtran}
\bibliography{IEEEexample}

\end{document}